\documentclass{l4dc2027}

\makeatletter
\g@addto@macro\ps@jmlrtps{%
  \let\@oddhead\@empty
  \let\@evenhead\@empty
}
\makeatother

\usepackage{times}
\usepackage{booktabs}
\usepackage{microtype}
\usepackage{placeins}
\usepackage{wrapfig}
\usepackage{needspace}
\usepackage[font=small]{caption}
\usepackage{xcolor} 

\usepackage[normalem]{ulem}

\RestyleAlgo{ruled}
\SetAlFnt{\small}
\SetAlCapFnt{\small}
\LinesNumbered
\SetCommentSty{textit}

\title[Recovery-Guided Model-Based Diffusion]{Recovery-Guided Model-Based Diffusion for Safety-Critical Legged Locomotion}

\author[Zheng, Sood, Vanniasinghe, Sartoretti and Lederer]{%
 \Name{Lei Zheng}$^{1}$ \Email{zack.zheng@nus.edu.sg}\\
 \Name{Shivam Sood}$^{2}$ \Email{shivamsood@u.nus.edu}\\
 \Name{Mithun Vanniasinghe}$^{1}$ \Email{mithun.vanniasinghe@u.nus.edu}\\
 \Name{Guillaume Sartoretti}$^{2}$ \Email{guillaume.sartoretti@nus.edu.sg}\\
 \Name{Armin Lederer}$^{1}$ \Email{armin.lederer@nus.edu.sg}\\
 \addr $^{1}$Department of Electrical and Computer Engineering\\
 \addr $^{2}$Department of Mechanical Engineering\\
 \addr National University of Singapore, Singapore
}

\begin{document}
\flushbottom

\maketitle 
\begin{abstract}%
Safety-critical high-speed legged locomotion requires motion planners to balance safety foresight with online computational efficiency. Although short prediction horizons reduce rollout costs, they provide limited safety foresight, and strict safety constraints severely restrict the feasible trajectory space. We propose a recovery-guided model-based diffusion (MBD) framework for safety-critical legged locomotion. It incorporates a learned terminal recovery critic into short-horizon trajectory optimization. Specifically, the critic approximates the worst-case future clearance margin along rollouts of a designated recovery policy. We introduce a feasibility-guided refinement strategy that combines within-horizon constraints and terminal recoverability at each sampling stage, while retaining feasible elites to mitigate sampling-support loss. Across 50 paired randomized Unitree Go2 obstacle-avoidance trials, our approach achieves zero collisions and 98\% strict-clearance success, while requiring substantially less computation than long-horizon MBD baselines.  A project page is available at \url{https://zack4417.github.io/rg-mbd/}.
\end{abstract}

 
\section{Introduction}
 Safety-critical legged locomotion in contact-rich environments requires controlling high-dimensional, nonlinear hybrid systems while satisfying stringent physical constraints~\citep{he2024abs, liao2023walking, sood2024decap}. The resulting trajectory optimization problem is highly nonconvex and admits distinct motion strategies, making diffusion models attractive for exploring multimodal trajectory distributions~\citep{janner2022diffuser,chi2023diffusionpolicy,zheng2026dynamic}. However, conventional diffusion-based planners typically rely on offline trajectory data, making performance sensitive to dataset coverage and quality. In addition, they represent system dynamics implicitly through data rather than explicitly enforcing them through
model rollouts. 

Recent methods address this limitation through incorporating explicit dynamic constraints~\citep{huang2026sadflower,yang2026uniconflow}, dynamics-aware trajectory scoring~\citep{li2024stabilizing}, or projection into reachable sets~\citep{bouvier2025ddat}. Nevertheless, maintaining strict dynamic feasibility remains challenging for high-dimensional legged locomotion. Model-based diffusion (MBD) circumvents these challenges by constructing generative updates from explicit model rollouts and task objectives~\citep{pan2024mbd}. Consequently, sampled trajectories remain dynamically consistent with the model without requiring task-specific diffusion training. 
 DIAL~\citep{xue2024dial} extends model-based generative refinement to online full-order legged locomotion using massively parallel model rollouts.  
However, finite-horizon planning does not ensure safety beyond the prediction horizon. While longer horizons provide greater foresight, they incur substantially higher computational costs.  In contrast, shorter horizons are computationally efficient but can produce trajectories that satisfy within-horizon constraints while terminating in states from which safe recovery is no longer possible~\citep{yin2025nsvimpc,yang2026safe}.  

 Addressing this horizon limitation raises two coupled challenges. First, the planners require scalable estimates of whether a terminal state admits a safe continuation. Learned reach--avoid and barrier values have been used for policy switching~\citep{he2024abs}, safety filtering~\citep{so2024train}, and enforcing beyond-horizon constraints in optimal control~\citep{yin2025nsvimpc}. 
Although these approaches extend safety foresight through learned safety values and constraints, they do not incorporate terminal recoverability into the successive generative refinement stages of MBD, where such guidance can actively shape trajectory distributions to preserve feasible candidate support during the diffusion process. More recently, \citet{yang2026dualshield} incorporate Hamilton--Jacobi (HJ) reachability values into diffusion guidance and reactive shielding. However, this grid-based HJ computation scales poorly with state dimension, limiting its direct {applicability to} full-order legged dynamics. 

 Second, incorporating terminal recoverability alongside within-horizon safety and physical constraints further restricts the feasible set of control sequences, potentially leaving too few feasible samples for reliable refinement under finite sampling budgets.  Recent constrained MBD methods improve sampling efficiency through progressive barrier enforcement~\citep{mishra2025ebmbd} and adaptive constraint handling with feasibility projection~\citep{he2026mdcoas}. However, these methods primarily focus on constraint satisfaction during refinement, without explicitly addressing how terminal recoverability constraints affect the preservation of feasible candidates across successive MBD stages.  Thus, a framework is needed to jointly address beyond-horizon recoverability and feasible-support preservation. 

In this paper, we propose a recovery-guided MBD framework that encodes long-horizon recovery information offline and exploits it during short-horizon MBD. A neural terminal recovery critic is trained from closed-loop rollouts under a designated recovery policy to approximate the worst-case future clearance margin along induced trajectories. During online planning, this critic is combined with within-horizon clearance and physical-feasibility checks to guide each diffusion-style refinement stage.  
The main contributions are:
\begin{itemize} 
\item We introduce a learned terminal recovery critic and exploit it as a terminal criterion in short-horizon MBD. This extends the effective safety foresight horizon without costly long-horizon online rollouts at each planning step during high-dimensional legged robot locomotion.
\item We develop a feasibility-guided MBD refinement strategy that integrates within-horizon constraints with terminal recoverability, using feasible-first weighting and safe-elite retention to mitigate feasible-support loss under finite sampling budgets.
\item We validate the proposed framework in simulation on the Unitree Go2 over 50 paired randomized obstacle-avoidance trials. The proposed 10-step planner achieves zero collisions and 98\% strict-clearance success, compared with a 58\% collision rate and 22\% strict-clearance success for 50-step DIAL, while requiring substantially less planning time.
\end{itemize}

\section{Preliminaries and Problem Statement} 
We model the legged robot with discrete-time dynamics \begin{equation}
\label{eq:system_dynamics}
    x_{t+1}=f(x_t,u_t),
\end{equation} 
where $x_t\in\mathcal X$ is the robot state (e.g., floating-base and joint coordinates and velocities) and $u_t\in\mathcal U$ is the admissible control input.

\subsection{MBD-Inspired Trajectory Optimization}
\label{subsec:dial}
MBD considers the problem of optimizing the finite-horizon control sequence $\mathbf U=(u_0,\ldots,u_{H-1})$,
subject to the system dynamics in \eqref{eq:system_dynamics}.
The objective is $\mathcal J(\mathbf U;x_0)
=\sum_{k=0}^{H-1}\ell(x_k,u_k)$,
where $\ell$ penalizes reference-tracking error,
body-posture deviation, and control effort. 
  Following MBD~\citep{pan2024mbd}, this cost induces the target distribution
\begin{equation}
    p_0(\mathbf U)
    \propto
    \exp\!\left(
        -\frac{\mathcal J(\mathbf U)}{\lambda}
    \right),
    \label{eq:mbd_target}
\end{equation}
where temperature $\lambda>0$ controls concentration around low-cost sequences. Model rollouts directly evaluate $\mathcal J(\mathbf U)$, making the target density available up to normalization without learning a score model. Direct sampling from $p_0$ is difficult for highly nonconvex objectives and sharply concentrated distributions. Diffusion instead considers Gaussian-smoothed distributions
\begin{equation}
    p_m(\mathbf U)
    =
    \left(
        p_0 * \mathcal N(0,\Sigma^m)
    \right)(\mathbf U),
    \qquad m=M,\ldots,1,
    \label{eq:dial_intermediate_dist}
\end{equation}
where $*$ denotes convolution and $\Sigma^m$ decreases across refinement stages, progressing from exploration at larger covariance to finer refinement toward $p_0$.  DIAL~\citep{xue2024dial} implements this reverse refinement online with Monte Carlo score estimates from parallel model rollouts. At stage $m$, perturbations $\mathbf W^j\sim\mathcal N(0,\Sigma^m)$, $j=1,\ldots,N_s$, yield
\begin{gather}
    \mathbf U^{m-1}
    =
    \mathbf U^m
    +
    \Sigma^m\nabla_{\mathbf U}
    \log p_m(\mathbf U^m)
    \approx
    \mathbf U^m
    +
    \sum_{j=1}^{N_s}w_j^m\mathbf W^j,
    \label{eq:dial_update}
\\
    w_j^m
    =
    \frac{
        \exp\!\left[
        -\widetilde{\mathcal J}
        (\mathbf U^m+\mathbf W^j)/\lambda
        \right]
    }{
        \sum_{\ell=1}^{N_s}
        \exp\!\left[
        -\widetilde{\mathcal J}
        (\mathbf U^m+\mathbf W^\ell)/\lambda
        \right]
    },
    \label{eq:dial_weights}
\end{gather}
where $\widetilde{\mathcal J} =(\mathcal J-\bar{\mathcal J})/\sigma_{\mathcal J}$ is the stage-wise standardized cost, with
 {$\bar{\mathcal J}$ and $\sigma_{\mathcal J}$ denoting the empirical mean and standard deviation} computed over the current
candidate batch, respectively.

Repeating~\eqref{eq:dial_update} with decreasing noise yields diffusion-inspired annealing. Receding-horizon operation applies the first optimized control and initializes the next cycle with the shifted sequence. The first cycle starts from zero control with $M_{\mathrm{init}}\ge M$ stages; subsequent cycles use $M$.

\subsection{Problem Statement}
Let $d_{\mathrm{body}}:\mathcal X\to\mathbb R$ denote the minimum signed
clearance between the robot body segments and the closest obstacle, with $d_{\mathrm{body}}<0$
indicating penetration, and let $d_{\mathrm{safe}}>0$ be a prescribed
clearance margin. The safety function $h:\mathcal{X}\rightarrow\mathbb{R}$ is defined as $h(x):=d_{\mathrm{body}}(x)-d_{\mathrm{safe}}$, with the corresponding safe set $\mathcal C_h:=\{x\in\mathcal X\mid h(x)\geq0\}\subseteq\mathcal X$. Thus, $h(x)\geq0$ indicates satisfaction of the prescribed margin.

\Needspace{18\baselineskip}
\begin{wrapfigure}{r}{0.42\textwidth} 
\centering
    \captionsetup{font=small,justification=raggedright,singlelinecheck=false}
     \includegraphics[width=0.98\linewidth]{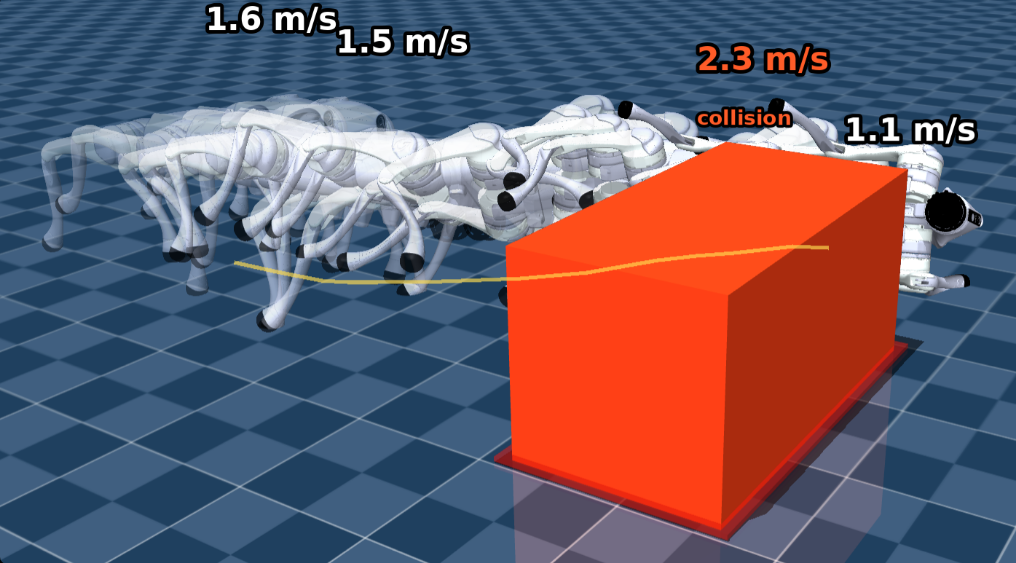}
    \captionof{figure}{Finite-horizon safety limitation with $H=10$. The yellow line shows the base trajectory. Although initially safe at $1.6\,\mathrm{m/s}$, the robot's terminal state fails the safety recovery criterion.}
    \label{fig:horizon_limitation} \vspace{-4mm}
\end{wrapfigure}   
We seek a control sequence that optimizes the cost 
$\min_{\substack{\mathbf U\in\mathcal U^H}}\mathcal J(\mathbf U;x_0)$,
subject to the dynamics~\eqref{eq:system_dynamics}
and $h(x_k)\geq0$ for $k=1,\ldots,H$. Real-time computational requirements restrict the  {prediction horizon $H$}. However, satisfying safety constraints over this horizon alone does not guarantee that the terminal state admits a safe continuation. We characterize this continuation using a designated admissible recovery policy $\pi_{\mathrm{rec}}$, defined in Sec.~\ref{sec:policy_recovery_value}. As Fig.~\ref{fig:horizon_limitation} illustrates, a trajectory can remain safe throughout the horizon yet terminate at a state that is not safely recoverable under $\pi_{\mathrm{rec}}$, ultimately leading to an unavoidable collision. Therefore, we seek trajectories that satisfy within-horizon constraints while terminating at states recoverable under this policy. 
 
\section{Methodology} 
\subsection{Terminal Recoverability}
\label{sec:policy_recovery_value}
Let $\pi_{\mathrm{rec}}:\mathcal X\rightarrow\mathcal U$ be a fixed deterministic recovery policy satisfying the control input limits, with induced closed-loop map $F_{\mathrm{rec}}(x) := f(x,\pi_{\mathrm{rec}}(x))$. Assume recovery trajectories are well-defined at every discrete step $k\geq0$ on a forward-invariant domain $\mathcal X_{\mathrm{op}}\subseteq\mathcal X$, with safety function $h$ bounded below along them.

From $x_0^{\pi_{\mathrm{rec}}}=x$, recovery follows $x_{k+1}^{\pi_{\mathrm{rec}}}=F_{\mathrm{rec}}(x_k^{\pi_{\mathrm{rec}}})$, with safety recovery value
\begin{equation}
    V^{h,\pi_{\mathrm{rec}}}(x)
    :=
    \inf_{k\in\mathbb N_0}
    h\!\left(x_k^{\pi_{\mathrm{rec}}}\right),
    \qquad x\in\mathcal X_{\mathrm{op}},
    \label{eq:recovery_value}
\end{equation}
the worst future safety margin under $\pi_{\mathrm{rec}}$. Following the policy-value approach to safety~\citep{yin2025nsvimpc}, this evaluates a prescribed continuation policy rather than optimizing over all policies.

The Bellman recursion is
\begin{equation}
    V^{h,\pi_{\mathrm{rec}}}(x)
    =
    \min\!\left\{
        h(x),
        V^{h,\pi_{\mathrm{rec}}}
        \!\left(F_{\mathrm{rec}}(x)\right)
    \right\}.
    \label{eq:recovery_bellman}
\end{equation}

The corresponding recovery set is
\begin{equation}
    \mathcal{C}_{\pi_{\mathrm{rec}}}
    :=
    \left\{
        {x\in\mathcal{X}_{\mathrm{op}}}
        \;\middle|\;
        V^{h,\pi_{\mathrm{rec}}}(x)\geq0
    \right\}.
    \label{eq:exact_recovery_set}
\end{equation}

\begin{proposition}
For nominal dynamics~\eqref{eq:system_dynamics}, the recovery set $\mathcal C_{\pi_{\mathrm{rec}}}$ satisfies $\mathcal C_{\pi_{\mathrm{rec}}}\subseteq\mathcal C_h$ and is forward invariant under $\pi_{\mathrm{rec}}$. For any constant $a\in(0,1)$ and linear function $\alpha(s)=as$,
\begin{equation}
    V^{h,\pi_{\mathrm{rec}}}(F_{\mathrm{rec}}(x))
    -
    V^{h,\pi_{\mathrm{rec}}}(x)
    \geq
    -\alpha\!\left(
        V^{h,\pi_{\mathrm{rec}}}(x)
    \right),
    \label{eq:exact_dcbf}
\end{equation}
where $x\in\mathcal C_{\pi_{\mathrm{rec}}}$.
\end{proposition}


\begin{proof}
 From the Bellman relation~\eqref{eq:recovery_bellman}, $V^{h,\pi_{\mathrm{rec}}}(x)\geq0$ implies $h(x)\geq0$, which yields $\mathcal C_{\pi_{\mathrm{rec}}}\subseteq\mathcal C_h$. Furthermore, \eqref{eq:recovery_bellman} gives $V^{h,\pi_{\mathrm{rec}}}(F_{\mathrm{rec}}(x))\allowbreak\geq V^{h,\pi_{\mathrm{rec}}}(x)\geq0$, which establishes  forward invariance. Since the left-hand side of \eqref{eq:exact_dcbf} is nonnegative and $-\alpha(V^{h,\pi_{\mathrm{rec}}}(x))\leq0$ on $\mathcal C_{\pi_{\mathrm{rec}}}$,  \eqref{eq:exact_dcbf} holds.
\end{proof} 

Thus, any terminal state in $\mathcal C_{\pi_{\mathrm{rec}}}$ admits a safe continuation ($h\geq0$) under $\pi_{\mathrm{rec}}$. We instantiate $\pi_{\mathrm{rec}}$ as an offline DIAL controller with horizon $H_{\mathrm{rec}}=50$, which is detailed in Sec.~\ref{sec:experimental_setup}.
\subsection{Neural Recovery Critic}
\label{sec:recovery_critic}
Inspired  by policy-value-based safety learning
\citep{so2024train}, we learn
$\mathcal B_\theta(z)$ as a practical approximation of
$V^{h,\pi_{\mathrm{rec}}}$ using temporal-difference learning
and finite-rollout supervision.  Here, $z=\phi(x)$ denotes a fixed feature
construction from robot-state and task-dependent
geometric quantities, as detailed in
Sec.~\ref{sec:task_safety}.

For recovery rollouts ending with successful obstacle passage at time $T_i$, the finite-rollout Monte Carlo (MC) target and one-step TD target are, respectively,
\begin{gather}
    b_t^{\mathrm{MC}}
    =
    \min_{t\leq k\leq T_i}h(x_k),
    \label{eq:critic_mc_target}
\\
    y_t
    =
    \min\!\left\{
        h(x_t),
        \mathcal{B}_{\theta}(z_{t+1})
    \right\},
    \label{eq:critic_target}
\end{gather}
where $z_{t+1}=\phi(x_{t+1})$. On the final transition, bootstrapping is replaced by the observed
terminal margin, yielding $y_{T_i-1}=\min\{h(x_{T_i-1}),h(x_{T_i})\}$.
Note that the MC targets satisfy  $b_t^{\mathrm{MC}}=\min\{h(x_t),b_{t+1}^{\mathrm{MC}}\}$   with   $b_{T_i}^{\mathrm{MC}}=h(x_{T_i})$. Thus, MC supervision uses the observed minimum-clearance
return anchored at termination, while TD bootstraps
the same stopped-rollout recursion.
 
To prevent optimistic approximation errors from inducing false-safe predictions, we employ the asymmetric loss function
\begin{equation}
    \ell_{\mathrm{fs}}(e)
    =
    \begin{cases}
        w_{\mathrm{fs}}e^2, & e>0,\\
        e^2,                & e\leq0,
    \end{cases}
    \qquad w_{\mathrm{fs}}>1,
    \label{eq:false_safe_loss}
\end{equation}
where $e$ is prediction error and overestimation ($e>0$) incurs a larger penalty. The training loss is
\begingroup
\small
\begin{equation}
\begin{aligned}
\mathcal{L}_{\mathcal B}(\theta)
={}&
\lambda_{\mathrm{TD}}
\mathbb{E}_{(x_t,x_{t+1})\sim\mathcal D_{\mathrm{def}}}
\!\left[
    \ell_{\mathrm{fs}}
    \!\left(
        \mathcal{B}_{\theta}(z_t)-y_t
    \right)
\right] +
\lambda_{\mathrm{MC}}
\mathbb{E}_{x_t\sim\mathcal D_{\mathrm{def}}}
\!\left[
    \ell_{\mathrm{fs}}
    \!\left(
        \mathcal{B}_{\theta}(z_t)-b_t^{\mathrm{MC}}
    \right)
\right]
\\
&+
\lambda_{\mathrm{env}}
\mathbb{E}_{x_t\sim\mathcal D_{\mathrm{rec}}}
\!\left[
    \left[
        \mathcal{B}_{\theta}(z_t)-h(x_t)
    \right]_{+}^{2}
\right] +
\lambda_{\mathrm{dec}}
\mathbb{E}_{(x_t,x_{t+1})\sim\mathcal D_{\mathrm{safe}}}
\!\left[
    \left[
        (1-\kappa)\mathcal{B}_{\theta}(z_t)
        -
        \mathcal{B}_{\theta}(z_{t+1})
    \right]_{+}^{2}
\right],
\end{aligned}
\label{eq:critic_loss}
\end{equation}
\endgroup 
where $[r]_+:=\max\{r,0\}$, $\mathcal D_{\mathrm{rec}}$ is the recovery-rollout dataset; $\mathcal D_{\mathrm{def}}\subseteq\mathcal D_{\mathrm{rec}}$
contains transitions from rollouts ending when $h(x_{T_i})<0$ 
or the robot successfully passes the obstacle;
and
$\mathcal D_{\mathrm{safe}}\subseteq\mathcal D_{\mathrm{def}}$
contains recovery-policy transitions whose next states are non-terminal with nonnegative finite-rollout margins $b_t^{\mathrm{MC}}\geq0$.
The first two terms of~\eqref{eq:critic_loss}
encourage TD consistency and observed-rollout fit,
respectively. The third term penalizes predictions above $h(x_t)$,
consistent with
$V^{h,\pi_{\mathrm{rec}}}(x)\leq h(x)$.
The final term softly encourages a DCBF-style inequality~\citep{zeng2021safety} analogous
to~\eqref{eq:exact_dcbf} 
\begin{equation}
    \mathcal{B}_{\theta}(z_{t+1})
    \geq
    (1-\kappa)\mathcal{B}_{\theta}(z_t),
    \qquad \kappa\in[0,1),
    \label{eq:critic_decay_condition}
\end{equation}
along safe recovery-policy transitions. This term serves strictly as a training regularizer and does not establish a DCBF condition for the learned critic.

During online planning, consistency with the instantaneous clearance margin is enforced through
\begin{equation}
    \widetilde{\mathcal{B}}_{\theta}(x)
    :=
    \min\!\left\{
        h(x),
        \mathcal{B}_{\theta}(\phi(x))
    \right\}.
    \label{eq:critic_envelope}
\end{equation}
Since $h(x)<0$ violates the prescribed margin, this envelope prevents critic approximation errors from erroneously classifying an instantaneously unsafe state as recoverable. The approximate recovery set is
\begin{equation}
    \widehat{\mathcal C}_{\pi_{\mathrm{rec}}}
    :=
    \left\{
        x\in\mathcal X
        \;\middle|\;
        \widetilde{\mathcal B}_{\theta}(x)\geq0
    \right\}.
    \label{eq:approx_recovery_set}
\end{equation} 

\begin{algorithm2e}[!htb]
\caption{Recovery-Guided Model-Based Diffusion}
\label{alg:rg_mbd}
\DontPrintSemicolon
\SetAlgoLined

\textbf{Offline: Neural recovery critic}\;
Generate recovery rollouts under the fixed policy $\pi_{\mathrm{rec}}$
and construct $\mathcal D_{\mathrm{rec}}$, $\mathcal D_{\mathrm{def}}$, and
$\mathcal D_{\mathrm{safe}}$\;
Construct the MC and TD targets using
\eqref{eq:critic_mc_target}--\eqref{eq:critic_target} and train
$\mathcal B_\theta$ by minimizing~\eqref{eq:critic_loss}\;
Freeze $\theta$ and construct $\widetilde{\mathcal B}_\theta$
using~\eqref{eq:critic_envelope}\;

\BlankLine
\textbf{Online: Recovery-guided diffusion}\;
Initialize the warm-start control sequence
$\mathbf U_{\mathrm{warm}}\gets\mathbf 0$\;
\For{each control cycle $t$}{
    Set $M_t\gets M_{\mathrm{init}}$ if $t=0$, otherwise $M_t\gets M$,
    and initialize $\mathbf U^{M_t}\gets\mathbf U_{\mathrm{warm}}$\;
    $\mathcal E\gets\emptyset$\;
    \For{$m=M_t,\ldots,1$}{ 
        Sample   perturbations $\mathbf W_m^j\sim\mathcal N(0,\Sigma^m)$
        and form candidates $\mathbf U_m^j=\mathbf U^m+\mathbf W_m^j$, $\forall  \mathbf U^m \in \mathcal{E}$ \;
        Roll out candidates through $f$ and evaluate
        $J_m^j$, $\widetilde J_m^j$, and $S_m^j$
        using~\eqref{eq:joint_feasibility}\;
        $\mathcal F_m\gets\{j\mid S_m^j\leq0\}$\;
        Compute recovery-guided weights $w_m^j$
        using~\eqref{eq:recovery_guided_weights}\;
        Update $\mathbf U^{m-1}\gets\sum_{j=1}^{N_s}w_m^j\mathbf U_m^j$\;
        Retain up to $E$ feasible elites in $\mathcal E$:
        $\mathcal E\gets
        \operatorname{Best}_E
        \bigl(\{(\mathbf U_m^j,J_m^j):j\in\mathcal F_m\}\bigr)$\;
        \; 
    }
    $j^\star\gets
    \begin{cases}
    \arg\min_{j\in\mathcal F_1}J_1^j,
        & \mathcal F_1\neq\emptyset,\\
    \arg\min_{1\leq j\leq N_s}S_1^j,
        & \text{otherwise}
    \end{cases}$\;

    Apply the first action $u_t\gets(\mathbf U_1^{j^\star})_0$
    and set $\mathbf U_{\mathrm{warm}}\gets\operatorname{shift}(\mathbf U_1^{j^\star})$\;
}
\end{algorithm2e}

\subsection{Recovery-Guided Diffusion}
\label{sec:guided_diffusion} 
We incorporate recovery information into each reverse-refinement stage of the annealed Gaussian proposal in Sec.~\ref{subsec:dial}, as summarized in
Algorithm~\ref{alg:rg_mbd}. At stage $m\in\{M,\ldots,1\}$, let $\tau_m^j=(x_{0:H}^{m,j},\mathbf U_m^j)$ be the rollout of particle $j\in\{1,\ldots,N_s\}$. Within-horizon clearance, physical feasibility, and terminal recoverability are evaluated through
\begin{equation}
    S_m^j
    :=
    \max\!\left\{
        \max_{1\leq k\leq H}[-h(x_k^{m,j})],
        g_{\mathrm{phys}}(\tau_m^j),
        -\widetilde{\mathcal B}_{\theta}(x_H^{m,j})
    \right\},
    \label{eq:joint_feasibility}
\end{equation}
where $g_{\mathrm{phys}}(\cdot)\leq0$ denotes satisfaction of input and whole-body physical constraints. The feasible-particle set $\mathcal F_m:=\{j\mid S_m^j\leq0\}$ satisfies nominal rollout constraints and the learned terminal test.

Each reverse-refinement stage conditions particle weights on the feasible set:
\begin{equation}
    w_m^j
    \propto
    \begin{cases}
    \mathbf 1_{\{j\in\mathcal F_m\}}
    \exp(-\widetilde{\mathcal J}_m^j/\lambda),
    & \mathcal F_m\neq\emptyset,\\[1mm]
    \exp(-S_m^j/T_s),
    & \text{otherwise}.
    \end{cases}
    \label{eq:recovery_guided_weights}
\end{equation}
with weights normalized over sampled particles.  When feasible particles exist ($\mathcal{F}_m \neq \emptyset$), they retain standard task-driven MBD weights, whereas an empty feasible set triggers a least-violation fallback governed by temperature $T_s$. In early high-noise stages, the framework explores recoverable candidates across a broader proposal distribution, while decreasing noise progressively concentrates refinement on task optimization.


To mitigate feasible-candidate depletion during stochastic refinement,  we retain $E$ feasible elites based on task cost and $S_m^j$, reinserting them into the next proposal according to the noise schedule in Sec.~\ref{subsec:dial}. Upon completing the reverse-refinement sequence, the planner applies the first control action of the lowest-cost feasible trajectory in $\mathcal{F}_1$;  if $\mathcal{F}_1 = \emptyset$, it defaults to the minimum-$S_1^j$ trajectory as a least-violation fallback. 

\section{Experiments} \label{sec:experiments} 
We evaluate recovery-guided MBD on a Unitree Go2 for high-speed obstacle avoidance, benchmarking collision avoidance, feasible sampling support, and computational efficiency.

\subsection{Simulation Setup}
\label{sec:experimental_setup}
\paragraph{Simulation and implementation.}
\label{subsubsec:simulation_setup}
We simulate the Unitree Go2 in GPU-parallel MuJoCo MJX on an NVIDIA RTX~5090, modeling a floating base with $12$ actuated joints. Control inputs follow the DIAL Go2 formulation~\citep{xue2024dial}, while candidate trajectories are evaluated under full-order rigid-body contact dynamics. Unless specified otherwise, all baseline and proposed configurations share identical task objectives, actuator limits, sampling budgets, initial states, and random seeds.

\begin{figure}[tb]
    \centering
    \includegraphics[width=0.49\linewidth,trim=1.44bp 229.77bp 1.44bp 1.44bp,clip]{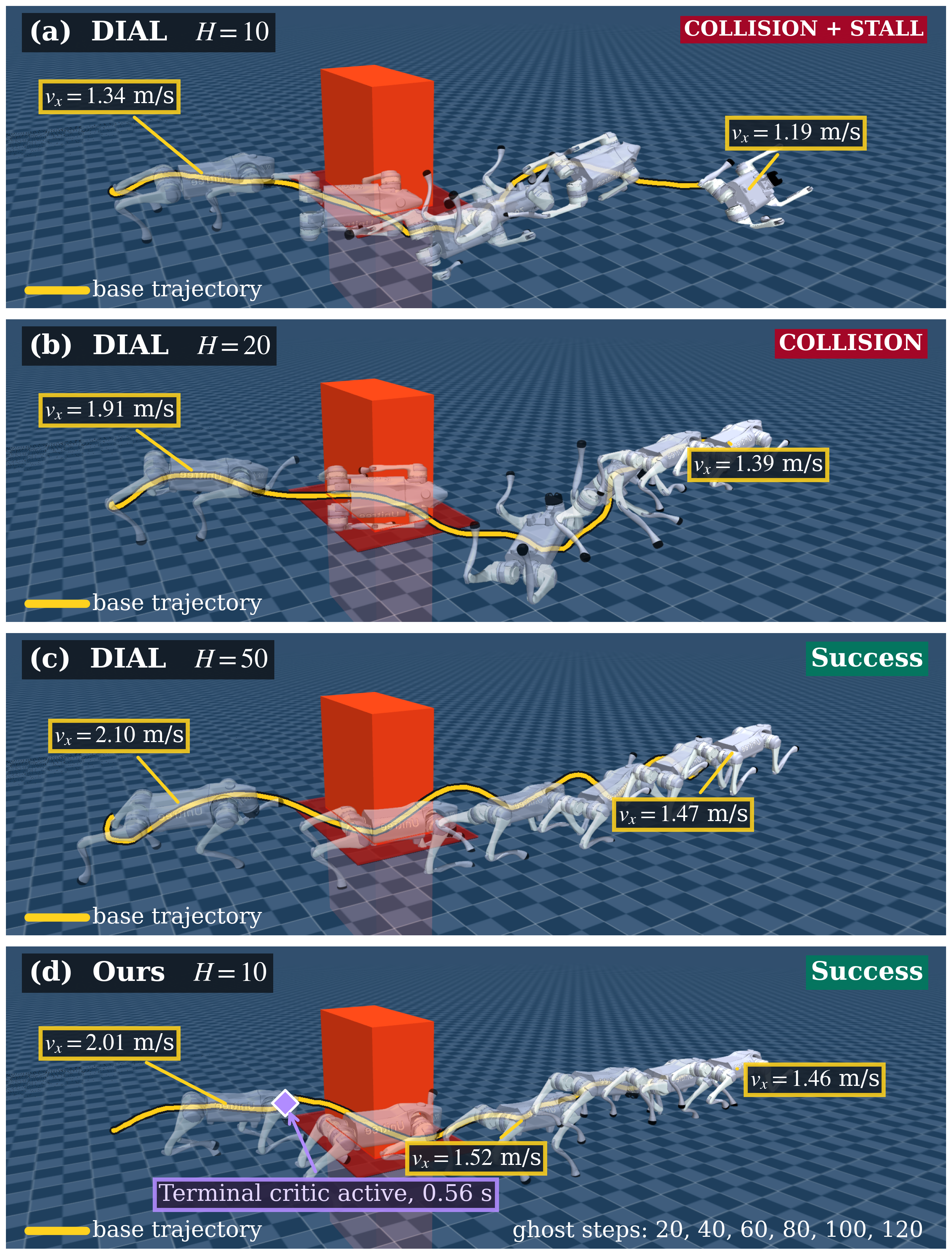}\hfill
    \includegraphics[width=0.49\linewidth,trim=1.44bp 153.63bp 1.44bp 77.49bp,clip]{Figures/fig_horizon_comparison_stroboscopic_2.png}
    \par\smallskip
    \includegraphics[width=0.49\linewidth,trim=1.44bp 77.58bp 1.44bp 153.63bp,clip]{Figures/fig_horizon_comparison_stroboscopic_2.png}\hfill
    \includegraphics[width=0.49\linewidth,trim=1.44bp 1.53bp 1.44bp 229.68bp,clip]{Figures/fig_horizon_comparison_stroboscopic_2.png}
    \caption{Unitree Go2 obstacle-avoidance snapshots: (a)--(c) DIAL with $H=10$, $20$, and $50$; (d) recovery-guided MBD with $H=10$. Yellow curves show the executed base trajectory, translucent poses show closed-loop states at 20-step intervals, and annotations give selected longitudinal base velocities.}
    \label{fig:sim_snap}\vspace{-2mm}
\end{figure}

\paragraph{Task and safety evaluation.}
\label{sec:task_safety}  
The robot tracks a $1.5\,\mathrm{m/s}$ forward reference velocity toward a $0.35\times0.70\times0.70\,\mathrm{m}$ ($\text{length}\times\text{width}\times\text{height}$) obstacle (Fig.~\ref{fig:sim_snap}).  
 Given $H=10$ and $\Delta t=0.02\,\mathrm{s}$, the $0.2\,\mathrm{s}$ horizon provides approximately $0.3\,\mathrm{m}$ of look-ahead at the commanded speed. Collision safety uses the minimum signed clearance $d_{\mathrm{body}}$ over the base, hips, thighs, and calves, excluding the feet. Strict safety requires $d_{\mathrm{body}}\ge2\,\mathrm{cm}$ throughout an episode; $d_{\mathrm{body}}<0$ denotes geometric penetration. Physical feasibility follows the  {fall, posture, joint, actuator, and harmful-contact conditions}
  in $g_{\mathrm{phys}}$.

\Needspace{17\baselineskip}
\begin{wraptable}{r}{0.52\textwidth}
    \vspace{-3mm}
    \centering
    \footnotesize
    \caption{Key implementation parameters.}
    \label{tab:implementation}
    \setlength{\tabcolsep}{3pt}
    \renewcommand{\arraystretch}{1.05}
    \begin{tabular}{@{}p{0.27\linewidth}p{\dimexpr0.73\linewidth-2\tabcolsep\relax}@{}}
    \toprule
    \textbf{Component} & \textbf{Parameter and value} \\
    \midrule
    Online planner
    & $H=10,\;\Delta t=0.02\,\mathrm{s}$\newline
      $N_s=2048,\;M=2,\;M_{\mathrm{init}}=6$ \\
    DIAL sampling
    & $\lambda=0.05,\;\rho_\sigma=0.5$\newline
      $\sigma_{\mathrm{ctl}}=[1,1,1]$ \\
    Safety guidance
    & $d_{\mathrm{safe}}=0.02\,\mathrm{m}$\newline
      $E=32,\;T_s=0.1$ \\
    Recovery policy
    & $H_{\mathrm{rec}}=50,\;N_{\mathrm{rec}}=2048$\newline
      $M_{\mathrm{rec}}=4,\;L_{\mathrm{rec}}=60$ steps \\
    Critic
    & $w_{\mathrm{fs}}=5,\;\kappa=0.05$ \\
    Critic loss
    & $\lambda_{\mathrm{TD}}=\lambda_{\mathrm{MC}}=1$\newline
      $\lambda_{\mathrm{env}}=0.25,\;\lambda_{\mathrm{dec}}=0.10$ \\
    Guidance\newline ablations
    & $\lambda_{\mathrm{soft}}=\lambda_{\mathrm{ds}}=10$\newline
      $a_{\mathrm{step}}=0.05$ \\
    \bottomrule
    \end{tabular}
    \vspace{-2mm}
\end{wraptable}
The critic $\mathcal B_\theta$ (Sec.~\ref{sec:recovery_critic}) uses $25$ standardized features $z=\phi(x)$: obstacle-relative position ($3$), box half-size ($3$), body-frame linear and angular velocities ($3{+}3$), projected gravity ($3$), base height ($1$), gait phase ($2$), commanded speed and target lane ($2$), and five grouped clearances ($5$, including the foot channel). Standardization uses the training split. The $25\! -\!128\! -\!128\! -\!128\! -\!1$ ReLU multilayer perceptron has a linear output and is trained offline on the stopped-rollout targets in Sec.~\ref{sec:recovery_critic}. A 
 DIAL controller with fixed settings and $H_{\mathrm{rec}}=50$ generates the recovery data.

 Recovery rollouts start at the terminal states of $H=10$ planning candidates.  Labels come from closed-loop execution of $\pi_{\mathrm{rec}}$, rather than imitation of task-oriented $H{=}50$ DIAL. Relabeling perturbs initial states with independent Gaussian noise: planar base position, $(\sigma_x,\sigma_y)=(0.08,0.10)\,\mathrm{m}$; base linear velocity, $\sigma_v=0.15\,\mathrm{m/s}$; joint position, $\sigma_q=0.025\,\mathrm{rad}$; and joint velocity, $\sigma_{\dot q}=0.05\,\mathrm{rad/s}$. Table~\ref{tab:implementation} lists the remaining parameters.

 A recovery rollout terminates at the earliest occurrence of one of four conditions: (i) a $d_{\mathrm{safe}}$ clearance violation (unsafe); (ii) obstacle passage without observed clearance or locomotion failure (successful observed recovery); (iii) locomotion failure; or (iv) timeout.  Ground contacts are logged but alone do not terminate rollouts.

\paragraph{Baselines and metrics.}
\label{sec:baselines_metrics}
We compare DIAL~\citep{xue2024dial} with $H\in\{10,20,50\}$ to assess the look-ahead, safety, and computation trade-off. Refinement budgets are $(M,M_{\mathrm{init}})=(2,6)$ for $H=10$ and $(4,10)$ for $H\in\{20,50\}$. Shield-DIAL applies a one-step DCBF-inspired filter following Shield-MPPI~\citep{yin2024shield}, restricting $u_0$ through $h(x)=d_{\mathrm{body}}(x)-d_{\mathrm{safe}}$ and $h(x^+)\ge(1-\alpha)h(x)$, with $\alpha=0.3$ and $d_{\mathrm{safe}}=2\,\mathrm{cm}$. Local search over $u_0$ perturbations repairs infeasible commands without changing DIAL's reverse-refinement loop. Our method uses the same $H=10$ DIAL configuration and temperature $\lambda$, conditioning each stage on $\mathcal F_m=\{j:S_m^j\leq0\}$.



To isolate the impact of the feasibility-guided weighting scheme, we evaluate two guidance ablations using the same $H=10$ planner and critic. \emph{Soft-guided} retains all candidate particles and applies soft constraint penalties via $\ell_m^j=-\widetilde{\mathcal J}_m^j/\lambda-\lambda_{\mathrm{soft}}[S_m^j]_+^2$. \emph{Dual-shield} maintains the hard feasibility gate $\mathcal F_m$, ranking feasible candidates via $\ell_m^j=-\widetilde{\mathcal J}_m^j/\lambda-\lambda_{\mathrm{ds}}S_m^j$ for $j\in\mathcal F_m$. When available, Soft-guided executes the task-optimal feasible candidate, defaulting to its highest-scoring candidate otherwise. Conversely, Dual-shield employs the least-violation fallback whenever $\mathcal F_m=\emptyset$.

We report collision rate $r_{\mathrm{col}}$, strict-$2\,\mathrm{cm}$ safety rate $r_{\mathrm{safe}}$, minimum whole-body clearance per episode $d_{\min}$, and mean feasible-particle fraction $\rho_{\mathrm{feas}}$ under $S_m^j\leq0$. Mean planning latency $t_{\mathrm{com}}$ and its 95th percentile $t_{\mathrm{p95}}$ exclude compilation and warm-up.


\subsection{Comparison with Baselines} 
Table~\ref{tab:go2_paired_randomized} summarizes $50$ paired randomized scenarios. DIAL collision rates are $98.0\%$, $68.0\%$, and $58.0\%$ for $H=10$, $20$, and $50$, respectively. Notably, no $H=10$ trial satisfies the strict $2\,\mathrm{cm}$ margin. As depicted in Fig.~\ref{fig:sim_snap}(a)--(c), short-horizon DIAL reacts too late to initiate evasive maneuvers, whereas extending the prediction horizon to $H=50$ enables partial obstacle avoidance and satisfies the margin in $22.0\%$ of trials.  Nevertheless, all three baseline configurations exhibit negative average minimum clearances. Thus, while extending the prediction horizon mitigates collisions, it incurs a severe computational penalty while leaving failure rates unacceptably high.

\Needspace{18\baselineskip}
\begin{wraptable}{r}{0.60\textwidth}
\vspace{-4mm}
    \centering
    \footnotesize
    \captionsetup{font=footnotesize,justification=raggedright,singlelinecheck=false}
    \caption{Quantitative comparison over $50$ paired randomized
obstacle-avoidance scenarios. Shield-DIAL augments DIAL with a
discrete CBF post-filter on $u_0$~\citep{yin2024shield}. Dual-shield
and Soft-guided are ablations. Bold denotes the best result.}
\label{tab:go2_paired_randomized}
\setlength{\tabcolsep}{1pt}
\renewcommand{\arraystretch}{1.02}
\begin{tabular*}{\linewidth}{@{\extracolsep{\fill}}lccccccc@{}}
\toprule
\textbf{Method} & $H$ &
$r_{\mathrm{col}}$ &
$d_{\min}$ &
$r_{\mathrm{safe}}$ &
$\rho_{\mathrm{feas}}$ &
$t_{\mathrm{com}}$ &
$t_{\mathrm{p95}}$ \\
& &
(\%) $\downarrow$ &
(cm) $\uparrow$ &
(\%) $\uparrow$ &
(\%) $\uparrow$ &
(ms) $\downarrow$ &
(ms) $\downarrow$ \\
\midrule
DIAL & $10$
& $98.0$
& $-2.77{\pm}2.50$
& $0.0$
& --
& $\mathbf{31.0{\pm}0.1}$
& $\mathbf{31.2}$ \\
DIAL & $20$
& $68.0$
& $-0.13{\pm}4.07$
& $20.0$
& --
& $113.9{\pm}0.5$
& $114.7$ \\
DIAL & $50$
& $58.0$
& $-0.37{\pm}3.52$
& $22.0$
& --
& $273.3{\pm}1.5$
& $276.1$ \\
\midrule
Shield-DIAL & $10$
& $96.0$
& $-1.78{\pm}1.74$
& $0.0$
& --
& $31.3{\pm}0.1$
& $31.4$ \\
Shield-DIAL & $20$
& $68.0$
& $-0.60{\pm}3.23$
& $20.0$
& --
& $114.2{\pm}0.3$
& $114.4$ \\
Shield-DIAL & $50$
& $60.0$
& $-0.57{\pm}3.23$
& $20.0$
& --
& $273.4{\pm}0.8$
& $274.7$ \\
\midrule
\textbf{Ours} & $10$
& $\mathbf{0.0}$
& $3.57{\pm}2.75$
& $\mathbf{98.0}$
& $54.6$
& $100.1{\pm}0.1$
& $100.3$ \\
Dual-shield & $10$
& $2.0$
& $\mathbf{4.55{\pm}3.58}$
& $94.0$
& $\mathbf{56.9}$
& $100.1{\pm}0.1$
& $100.3$ \\
Soft-guided & $10$
& $20.0$
& $2.14{\pm}4.44$
& $80.0$
& $49.2$
& $100.0{\pm}0.1$
& $100.3$ \\
\bottomrule
\end{tabular*}\vspace{-2mm}
\end{wraptable}

Augmenting DIAL with a single-step control barrier filter yields marginal safety gains. At $H=10$, Shield-DIAL slightly reduces the collision rate from $98.0\%$ to $96.0\%$, while strict-$2\,\mathrm{cm}$ safety remains $0.0\%$. At $H=20$ and $H=50$, Shield-DIAL yields collision rates of $68.0\%$ and $60.0\%$, respectively, with lower mean clearance than plain DIAL. This confirms that instantaneous $u_0$ filtering cannot provide long-horizon safety under contact-rich dynamics. In contrast, recovery-guided MBD ($H=10$) achieves zero collisions, $98.0\%$ strict safety, and a positive mean clearance of $3.57\pm2.75\,\mathrm{cm}$. Crucially, its mean planning latency ($100.1\,\mathrm{ms}$) is below DIAL at $H=20$ ($113.9\,\mathrm{ms}$) and $H=50$ ($273.3\,\mathrm{ms}$), delivering superior safety at lower computational cost. 

\Needspace{18\baselineskip}
\begin{wrapfigure}{r}{0.54\textwidth}
\vspace{-4mm}
\centering
    \captionsetup{font=small,justification=raggedright,singlelinecheck=false}
    \includegraphics[width=0.98\linewidth]{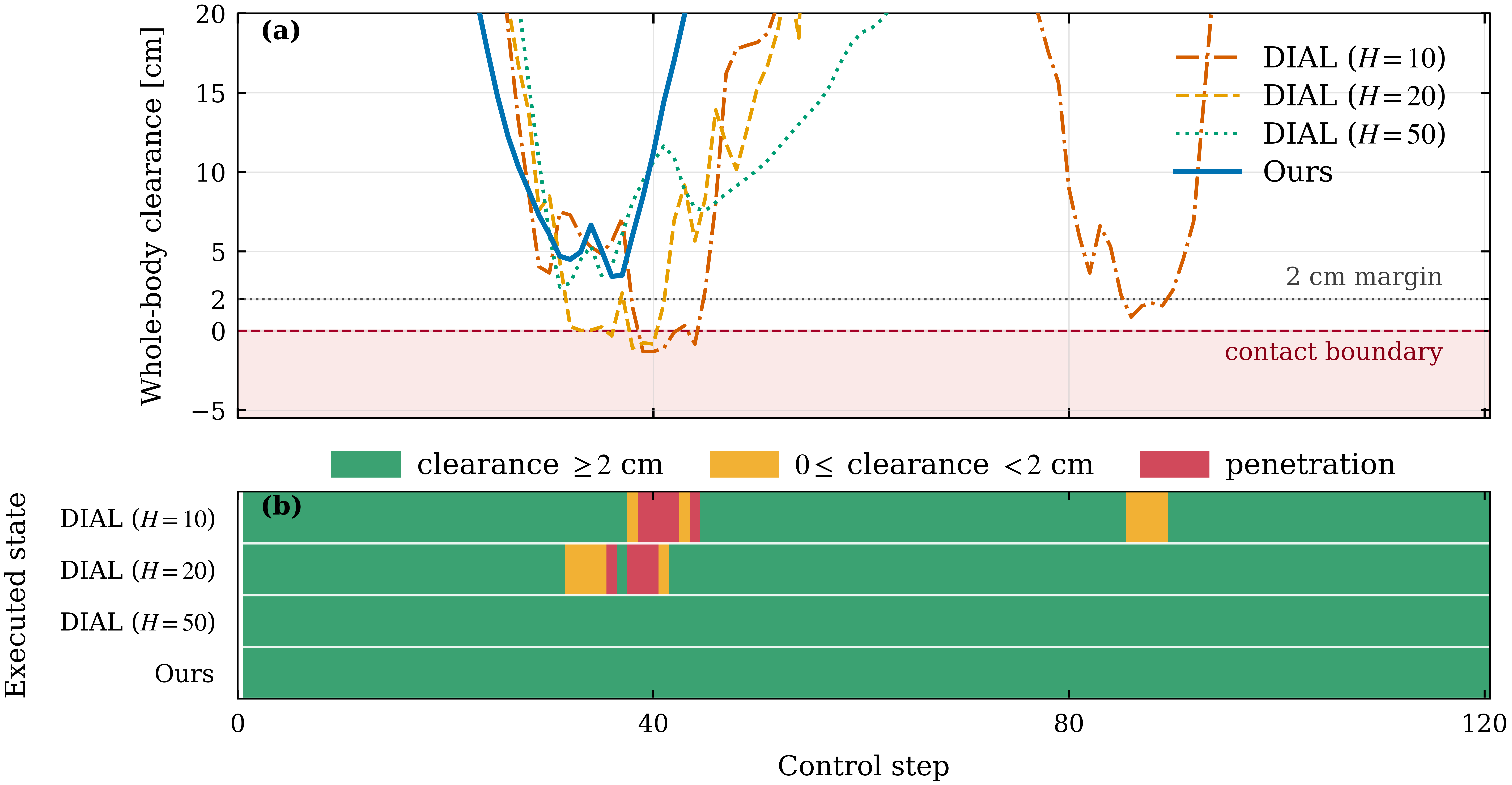}
    \caption{Safety comparison of DIAL ($H=10,20,50$) and recovery-guided MBD ($H=10$). \textbf{(a)} Minimum whole-body clearance; dotted and dashed lines mark the $2\,\mathrm{cm}$ safety margin and $0\,\mathrm{cm}$ contact boundary. \textbf{(b)} Executed safety state.}
    \label{fig:sim_safety} \vspace{-4mm}
\end{wrapfigure} 

Ablations highlight the role of hard feasibility enforcement during refinement. Dual-shield increases mean minimum clearance to $4.55{\pm}3.58\,\mathrm{cm}$ and feasible particle fraction to $56.9\%$, but slightly reduces strict safety to $94.0\%$. This secondary ranking improves candidate feasibility without adding latency. Conversely, Soft-guided results in $20.0\%$ collisions and $80.0\%$ safety, showing that soft penalty terms allow unsafe samples to compromise refinement compared to hard feasibility gating. 

Figures~\ref{fig:sim_snap} and~\ref{fig:sim_safety} show representative closed-loop rollouts. DIAL with $H=10$ and $20$ reacts too late and intersects the obstacle; $H=50$ avoids it but approaches the $2\,\mathrm{cm}$ margin. Recovery-guided MBD with $H=10$ starts avoidance earlier and maintains the margin. DIAL's aggregate $58.0\%$ collision rate at $H=50$ shows that this illustrated benefit is inconsistent across randomized scenarios, despite the longer look-ahead. 
\subsection{ {Sampling Budget and Runtime}}
\label{subsec:budget_ablation}
 {To evaluate real-time viability, we sweep the sample budget $N_s$ and refinement stages $M$ on a representative high-speed maneuver.} All other settings follow Sec.~\ref{subsubsec:simulation_setup}. 
\Needspace{15\baselineskip}
\begin{wrapfigure}{r}{0.54\textwidth}
\vspace{-2mm}
\centering
    \captionsetup{font=small,justification=raggedright,singlelinecheck=false}
    \includegraphics[width=0.98\linewidth]{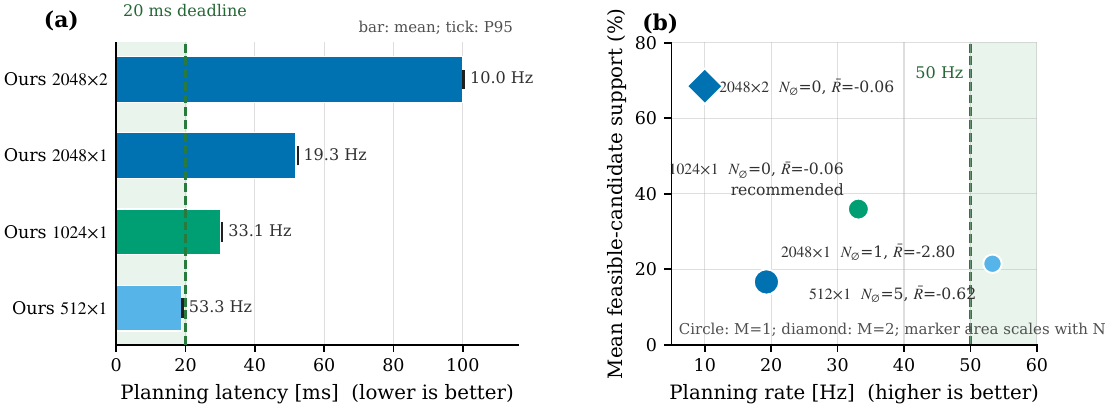}
    \caption{Computational-budget ablation. \textbf{(a)} Mean latency; black ticks show 95th percentiles and shading marks the $20\,\mathrm{ms}$ ($50\,\mathrm{Hz}$) real-time deadline. \textbf{(b)} Planning frequency versus feasible-candidate support; $N_{\emptyset}$ counts empty-support steps and $\bar R$ is the mean reward.}
    \label{fig:sim_traj_ablation}
\vspace{-2mm}
\end{wrapfigure} At $N_s=2048$, reducing $M$ from two to one nearly doubles the planning frequency, as shown in Fig.~\ref{fig:sim_traj_ablation}.  With $N_s=1024$ and $M=1$, the planner reaches $33.13\,\mathrm{Hz}$ with continuous feasible support ($N_{\emptyset}=0$) and the largest minimum clearance tested. 
Reducing $N_s$ to $512$ achieves a mean planning frequency
of $53.32\,\mathrm{Hz}$ ($18.75\,\mathrm{ms}$ mean latency),
with five empty-support steps and reduced task performance.
All configurations remain collision-free in this maneuver;
$N_s=1024,M=1$ balances task performance and feasible support,
while $N_s=512,M=1$ meets the $20\,\mathrm{ms}$ planning
budget on average. 

\subsection{Terminal versus Stepwise Recovery Constraints}
\label{subsec:stepwise_ablation}
We ablate whether adding intermediate DCBF-style constraints to \emph{Ours}
improves over recoverability filtering alone. The variant keeps the same
learned critic but additionally requires $\widetilde{\mathcal B}_\theta(x_{k+1}) \geq (1-a_{\mathrm{step}})\widetilde{\mathcal B}_\theta(x_k)$ for all $k \in \{0, \dots, H-1\}$  along each DIAL-predicted candidate
(not closed-loop $\pi_{\mathrm{rec}}$ transitions~\eqref{eq:exact_dcbf}). A dedicated benchmark of 50 paired randomized runs with fixed obstacle geometry isolates constraint-induced support loss.

\Needspace{10\baselineskip}
\begin{wraptable}{r}{0.47\textwidth}
    \vspace{-3mm}
    \centering
    \footnotesize
    \caption{Paired ablation of recoverability filtering
    (\emph{Ours}) versus the same critic with added stepwise
    constraints across 50 paired runs under fixed obstacle geometry.}
    \label{tab:stepwise_ablation}
    \setlength{\tabcolsep}{3pt}
    \renewcommand{\arraystretch}{1.05}
    \begin{tabular}{@{}lcc@{}}
        \toprule
       \textbf{Metric} & \textbf{Ours} &    \shortstack{\textbf{Ours +}\\\textbf{stepwise}} \\
        \midrule
        Collision-free episodes & 100\% & 94\%  \\
        Task success & 94\%  & 66\% \\
        Feasible candidate fraction & 71.72\% & 31.22\% \\
        Empty-support steps ($N_{\emptyset}$) & 4 & 493 \\
        Mean min.\ clearance & 2.49 cm & 21.10 cm \\
        Mean reward & $-0.0586$ & $-1.0248$ \\
        \bottomrule
    \end{tabular}
\vspace{-2mm}
\end{wraptable}

In Table~\ref{tab:stepwise_ablation}, \emph{Ours} achieve 100\% collision-free episodes and 94\% task success. Stepwise enforcement reduces these to 94\% and 66\%, respectively, lowers the feasible-candidate fraction from $71.72\%$ to $31.22\%$, and increases empty-support steps from 4 to 493. Of these 493 steps, 154 contained terminally recoverable candidates before stepwise filtering, demonstrating that intermediate constraints directly trigger support collapse. Although mean minimum clearance increases to $21.10\,\mathrm{cm}$, this conservatism severely degrades task reward and induces collisions when support collapses.  \emph{Ours} thus preserve feasible support while maintaining high safety and task performance.

\section{Conclusion} 
In this work, we present a recovery-guided MBD framework for safety-critical legged locomotion. By integrating a policy-conditioned terminal recovery critic with feasibility-driven annealing and support retention, our approach maintains beyond-horizon recoverability and viable trajectory support.  Simulations on the Unitree Go2 show reduced collision rates relative to DIAL and Shield-DIAL baselines at lower computational cost than long-horizon planning.  Future work will extend the feasibility-guided support mechanism to uneven terrain under model uncertainty.
\subsection*{Generative AI Use}  
Generative AI tools were used for translation, language editing,
LaTeX formatting, and feedback on the technical presentation of the
manuscript. The authors
reviewed and verified all AI-assisted content and take responsibility for the
final content of the paper.


\bibliography{l4dc2027-sample}

\end{document}